# DanceFusion: A Spatio-Temporal Skeleton Diffusion Transformer for Audio-Driven Dance Motion Reconstruction

Li Zhao[*], Zhengmin Lu
Tsinghua University
zhaoli@tsinghua.edu.cn

## Abstract

*This paper introduces DanceFusion, a novel framework for reconstructing and generating dance movements synchronized to music, utilizing a Spatio-Temporal Skeleton Diffusion Transformer. The framework adeptly handles incomplete and noisy skeletal data common in short-form dance videos on social media platforms like TikTok. DanceFusion incorporates a hierarchical Transformer-based Variational Autoencoder (VAE) integrated with a diffusion model, significantly enhancing motion realism and accuracy. Our approach introduces sophisticated masking techniques and a unique iterative diffusion process that refines the motion sequences, ensuring high fidelity in both motion generation and synchronization with accompanying audio cues. Comprehensive evaluations demonstrate that DanceFusion surpasses existing methods, providing state-of-the-art performance in generating dynamic, realistic, and stylistically diverse dance motions. Potential applications of this framework extend to content creation, virtual reality, and interactive entertainment, promising substantial advancements in automated dance generation. Visit our project page at https://th-mlab.github.io/DanceFusion/.*

## 1. Introduction

### 1.1. Background

In the rapidly evolving landscape of social media, platforms like TikTok have revolutionized the creation, sharing, and global consumption of dance culture, transforming how audiences engage with and participate in performance art. These platforms popularize short, musically synchronized dance routines that captivate viewers with rhythmic precision and stylistic diversity. However, the spontaneous, user-generated nature of these performances often results in video data that is noisy, incomplete, and captured from suboptimal perspectives, posing significant challenges for computer vision and pose estimation models[1][2]. Studies have demonstrated the complexity of processing such unstructured data, emphasizing the need for models capable of handling occlusions, irregular viewpoints, and low-resolution visuals [3][4].

The analysis and synthesis of human motion have been pivotal research areas within the field of computer vision, with far-reaching applications in animation, gaming, human-computer interaction, and robotics[5][6][7][8][9]. The emergence of platforms like TikTok has introduced a novel dataset that diverges markedly from the controlled, high-fidelity datasets traditionally utilized in academic research. TikTok dances are often recorded in uncontrolled environments, yielding data that is frequently noisy, incomplete, or captured from suboptimal perspectives. Moreover, the tight coupling of these dances with their accompanying music necessitates models capable of not only accurately reconstructing or generating motion sequences but also ensuring that these sequences are rhythmically synchronized with the audio[10][11][12].

### 1.2. Technological Challenges

Traditional motion capture technologies are designed for controlled environments and require sophisticated equipment. Such conditions are unattainable with casual recordings from platforms like TikTok, resulting in significant data processing challenges. Current methods struggle to handle the unstructured data from these platforms effectively, suffering from issues like background noise and partial occlusions, which leads to unsatisfactory motion synchronization with background music.

### 1.3. Research Motivation

This study introduces DanceFusion, a framework that employs a Spatio-Temporal Skeleton Diffusion Transformer designed to address the challenges posed by social media dance videos. DanceFusion aims to improve the accuracy and realism of dance motion reconstructions from noisy and incomplete data, enhancing content creation for virtual reality and interactive media.

### 1.4. Research Objectives

The primary research objectives of this study are outlined as follows:
1. **Hierarchical Spatio-Temporal VAE for Motion Reconstruction**: To engineer a Spatio-Temporal Transformer-based Variational Autoencoder (VAE) that can accurately reconstruct incomplete TikTok dance motions, even amidst considerable data gaps or noise.
2. **Integration of Diffusion Models for Audio-Driven Motion Generation**: To integrate a diffusion model within the VAE framework to facilitate the generation of high-fidelity, audio-synchronized dance motions.
3. **Evaluation of Framework Efficacy**: To assess the efficacy of the proposed framework using a diverse dataset of TikTok dance sequences, evaluating both its reconstruction fidelity and its capacity to generate realistic, musically synchronized motion

### 1.5. Contributions

This study makes the following key contributions:
1. **Hierarchical Spatio-Temporal VAE for Motion Reconstruction**: We introduce a hierarchical Transformer-based Variational Autoencoder (VAE) that integrates spatio-temporal encoding to effectively capture both spatial joint configurations and temporal movement dynamics. This integration enhances the model's ability to reconstruct incomplete and noisy skeleton data with high accuracy.
2. **Integration of Diffusion Models for Audio-Driven Motion Generation**: By incorporating diffusion models, DanceFusion iteratively refines motion sequences, significantly improving motion realism and ensuring precise synchronization with audio inputs.
3. **Advanced Masking Techniques**: We develop sophisticated masking strategies to manage missing or unreliable joint data, allowing the model to prioritize and accurately reconstruct available information.

### 1.6. Paper Organization

The remainder of this paper is organized as follows:
- Chapter 2: Related Work reviews existing literature on motion generation and reconstruction, the application of VAEs and Transformers in motion analysis, diffusion models for audio-driven motion generation, and the utilization of TikTok dance data in machine learning.
- Chapter 3: Methodology details the architecture of the DanceFusion framework, including the hierarchical Transformer VAE, diffusion process, and handling of incomplete data.
- Chapter 4: Experimental Setup describes the dataset, evaluation metrics, and implementation specifics used to validate the framework.
- Chapter 5: Results and Discussion presents the experimental findings, comparing the proposed method with baseline approaches and discussing its effectiveness.
- Chapter 6: Conclusion and Future Work summarizes the key findings, discusses limitations, and outlines potential directions for future research.

## 2. Related Work

### 2.1. Skeleton-Based Action Recognition and Pose Estimation

Skeleton-based action recognition and pose estimation play a critical role in understanding human motion, with applications across surveillance, sports analytics, and human-computer interaction. Traditional methods that rely on handcrafted features struggle to handle the unstructured and complex data often found on social media. Recently, Graph Convolutional Networks (GCNs) such as AS-GCN[13], 2s-AGCN[14], and MSG3D[15] have advanced skeleton-based models by dynamically capturing spatial-temporal dependencies between joints, though they remain sensitive to occlusions and missing data[16]. To address these challenges, Transformer models have been introduced, leveraging self-attention to capture long-range dependencies and improve robustness, as seen in models like Skeletor[17]. Multi-modal approaches like PoseC3D[18], which fuse skeleton and RGB data, further enhance performance in handling noisy or incomplete inputs, making these models more adaptable to varying data quality.

### 2.2. Motion Generation and Reconstruction

Human motion generation and reconstruction are critical fields in computer vision and graphics, impacting animation, gaming, human-computer interaction, and robotics. Traditional methods, primarily reliant on motion capture systems, excel in accuracy but falter with the spontaneous, unstructured data from platforms like TikTok, which are ill-suited for such rigid systems.

Transformers have recently been introduced to motion analysis due to their ability to handle long-range dependencies and capture complex temporal dynamics. Models like MDM[7], TEMOS[19], MotionClip[5], MotionGPT[20][21] have explored translating textual descriptions into motion sequences. Despite these advances, effectively combining spatial and temporal encoding while managing incomplete or noisy input data remains a significant challenge. DanceFusion addresses this gap by integrating a hierarchical Transformer architecture with advanced masking techniques to handle incomplete skeleton data.

## 2.3. Variational Autoencoders in Motion Analysis

Variational Autoencoders (VAEs)[22] have become a popular tool for motion analysis due to their ability to learn compact and interpretable latent representations of complex data distributions. In the context of motion generation, VAEs have been used to encode human poses into a latent space from which new poses or sequences can be sampled. This capability is particularly useful for tasks that require the generation of diverse and realistic motion patterns.

Previous work has demonstrated the effectiveness of VAEs in tasks such as human pose estimation, where they are used to reconstruct missing joints or generate plausible pose sequences from partial observations. For example, the MLD[23] designed a VAE to capture a representative low-dimensional latent feature for human motion sequences, and ACTOR[24] trained a generative VAE to learn action-aware human motion latent representations, enabling the synthesis of variable-length motion sequences conditioned on action categories.

However, standard VAE models often face challenges when dealing with the high variability and temporal coherence required in motion data, particularly when the input data is incomplete or noisy. The DanceFusion framework builds on this foundation by integrating a hierarchical Transformer structure with the VAE, allowing for more sophisticated spatial and temporal encoding. This approach not only enhances the model's ability to reconstruct complex motion patterns but also improves its robustness to missing data by leveraging the flexibility of VAEs in handling uncertainty in the input data.

## 2.4. Transformer Models for Sequential Data

Transformers have revolutionized the processing of sequential data, especially in Natural Language Processing (NLP), with their self-attention mechanism that effectively captures long-range dependencies without relying on recurrence[25]. In computer vision, Transformers have been applied to tasks such as action recognition, motion prediction, and pose estimation by treating motion sequences as tokenized inputs similar to words in a sentence[26][27].

In motion analysis, Transformers have been used to model temporal dynamics and predict future motions. However, many of these models assume complete and well-structured input data, limiting their applicability to real-world scenarios with incomplete or corrupted data[28].

DanceFusion overcomes these limitations by employing a dual-level Transformer architecture within a VAE framework. This design simultaneously handles spatial and temporal encoding while effectively managing incomplete skeleton data through masking. The approach allows for the generation of realistic and coherent motion sequences from sparse or noisy inputs.

## 2.5. Diffusion Models in Audio-Driven Motion Generation

Diffusion models have emerged as powerful generative techniques for high-dimensional data such as images and audio[29]. They transform a simple distribution into a complex one resembling the target distribution through a series of learned denoising steps. In motion generation, diffusion models have been applied to generate realistic and synchronized motion sequences from text or audio inputs[23][30][31].

For example, the FineDance[11], EDGE[9] models have demonstrated outstanding performance in generating human dance motions using diffusion techniques. These models progressively refine motion sequences to match the rhythm and dynamics of the input music.

While diffusion models show promise, integrating them with other generative models like VAEs has been less explored. DanceFusion pioneers this integration by combining a VAE with a diffusion model for audio-driven motion generation. This fusion enables the model to produce synchronized and stylistically appropriate dance sequences that reflect both the audio characteristics and the temporal structure of the motion data.

## 2.6. TikTok Dance Data in Machine Learning

TikTok's rise has unveiled a rich vein of dance motion data, marked by short, rhythmic sequences that sync with popular music, presenting unique challenges and opportunities in machine learning. This data presents unique challenges for machine learning due to its unstructured nature, variability in performance quality, and frequent presence of incomplete or noisy skeleton data.

Recent research has explored generating realistic human dance videos by leveraging both images and skeletal representations. Methods such as DISCO[32], Animate Anyone[33], MagicAnimate[34], and MagicPose[35] utilize advanced techniques like ControlNet to condition image generation on skeletal inputs. These approaches effectively generate visually coherent and temporally consistent dance videos by controlling image synthesis through skeleton graphs. These projects highlight the potential of combining skeletal motion data with image synthesis techniques to produce engaging dance content. However, the performance of these methods heavily relies on the quality and completeness of the input skeletal sequences, as almost all skeletal sequences are derived from pre-existing video footage.

**DanceFusion** serves as an upstream solution to these approaches by generating high-quality, realistic skeleton sequences from partial or corrupted data. The sequences produced by DanceFusion can be used as inputs to these image-based generation methods, enhancing their performance and enabling the creation of more accurate and synchronized dance videos. By providing robust and refined skeleton sequences, DanceFusion contributes to the overall pipeline of dance video generation, bridging the gap between motion reconstruction and high-fidelity visual synthesis.

In essence, DanceFusion not only addresses the challenges associated with incomplete skeleton data but also augments subsequent stages of dance video generation. Its ability to produce reliable skeletal motion sequences makes it a valuable component that can be integrated with existing models for image and video synthesis, thereby expanding the possibilities for creating high-quality, synchronized dance content on social media platforms.

## 3. Methodology: Spatio-Temporal Skeleton Diffusion Transformer

### 3.1. Overview of the DanceFusion Framework

In this chapter, we present the DanceFusion framework, which integrates a hierarchical **Spatio-Temporal Transformer-based Variational Autoencoder (VAE)** with a diffusion model to achieve robust motion reconstruction and audio-driven dance motion generation. The input to the model is a sequence of skeleton joints extracted from TikTok dance videos, often containing missing or noisy data. The framework aims to refine these skeleton sequences through a series of denoising steps, ensuring that the generated or reconstructed motion is temporally consistent and aligned with accompanying audio.

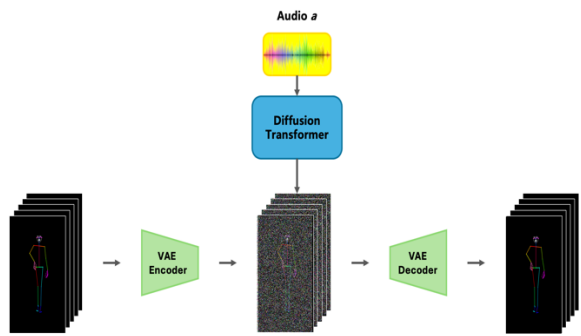

Figure 1: Overview of DanceFusion Framework.

The framework operates in three key phases:
- **Hierarchical Spatio-Temporal VAE Encoding**: The framework employs a hierarchical Transformer-based VAE to capture both the spatial configurations of joints and the temporal dynamics of movements. This integration enhances the model's ability to reconstruct incomplete and noisy skeletal data with high accuracy.
- **Diffusion Process for Motion Refinement**: By incorporating a diffusion model within the VAE framework, DanceFusion iteratively refines motion sequences, significantly improving motion realism and ensuring synchronization with audio inputs.
- **Advanced Masking Techniques**: We introduce masking strategies to handle missing or unreliable joint data, allowing the model to focus on available information and improve reconstruction accuracy.

### 3.2. Hierarchical Spatio-Temporal VAE Encoding

The DanceFusion framework introduces a hierarchical Transformer-based Variational Autoencoder (VAE) that integrates spatio-temporal encoding to capture the spatial and temporal information inherent in skeleton sequences. Unlike image-based models like ViViT[26], which processes grid patches from static images, we treat each skeleton joint as a token, and the sequence of joints over time forms a spatio-temporal grid. Each joint in the skeleton is encoded based on its spatial relationship with other joints, and the temporal sequence of these joint positions is fed into the Transformer for processing.

A key challenge in this process is handling missing joints due to occlusions or sensor noise. To address this, we introduce a masking mechanism during the embedding process. Each joint position is embedded as a high-dimensional vector using linear layers, but for missing joints, a masking signal is applied to prevent these missing values from influencing the encoding. This allows the model to focus on the available data while effectively ignoring unreliable or missing input.

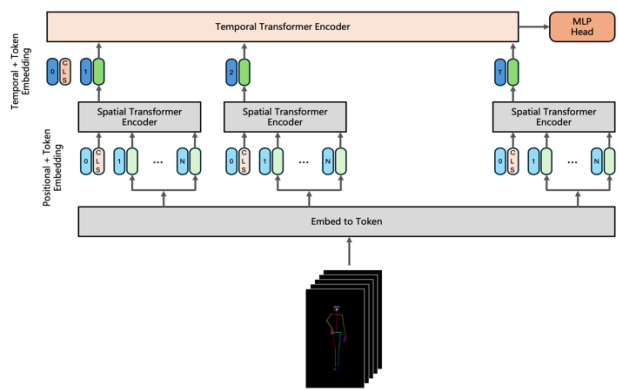

Figure 2: Spatio-Temporal Encoding.

#### 3.2.1 Embedding Skeletal Data

The input to the Spatio-Temporal Encoding module is a sequence of 2D or 3D skeletal joints. Each frame is represented as a set of joint coordinates, which are then embedded into a higher-dimensional space. This embedding process is similar to the patch embedding used in ViViT[26] but tailored to skeletal data. Specifically, we map the coordinates of each joint into a token that encapsulates both its spatial location and its reliability (in the case of missing data). This embedding allows us to apply transformer layers that can capture complex dependencies across both space and time.

#### 3.2.2 Spatial Transformer Encoder

After embedding the joint coordinates into tokens, we apply a Spatial Transformer Encoder to model the spatial relationships within each frame. This step is critical because it allows the model to understand the configuration of the joints, even in the presence of missing data. The spatial transformer operates on the tokens corresponding to joints in the same frame, capturing local and global spatial patterns.

For frames with missing joints, the corresponding tokens are masked out, preventing the model from relying on incomplete data during this stage. This masking ensures that the spatial transformer focuses on the available and reliable data, thereby improving the robustness of the model.

#### 3.2.3 Temporal Transformer Encoder

Following the spatial encoding, we apply a Temporal Transformer Encoder to capture the temporal dynamics of the skeletal motion. The temporal transformer processes the sequence of spatially encoded tokens, learning the temporal dependencies between frames. This encoder is particularly important for understanding the continuity of motion and the progression of the dance sequence over time.

The temporal transformer ensures that the model can generate smooth and coherent motion sequences, even when some frames contain incomplete data. By focusing on the temporal structure, the model compensates for any missing joints by relying on the temporal context provided by the surrounding frames.

#### 3.2.4 Output Representation

The output of the Spatio-Temporal Encoding module is a sequence of encoded tokens that capture both the spatial configuration of joints within each frame and the temporal progression of these configurations across frames. This representation serves as the input to subsequent stages of the DanceFusion framework, where it is further processed to generate refined and realistic dance motions.

This approach effectively combines the strengths of ViViT's transformer architecture with adaptations tailored to the unique challenges posed by skeletal data, ensuring that the model can handle incomplete input data while still generating high-quality motion sequences.

### 3.3. Loss Function Design

The overall loss function used in the DanceFusion framework is designed to balance motion reconstruction accuracy and latent space regularization. We employ a combination of the reconstruction loss and the Kullback-Leibler (KL) divergence loss to ensure that the model learns a compact and interpretable latent representation while accurately reconstructing skeleton sequences.

The total loss function $L_{total}$ is defined as a weighted sum of the reconstruction loss (either MSE or L1) and the Kullback-Leibler (KL) divergence loss:

$$L_{total} = L_{recon} + \beta \cdot L_{KL} \quad (1)$$

Where:
- $L_{recon}$ is either the MSE Loss or the L1 Loss, depending on the specific experiment.
- $\beta$ is a hyperparameter that controls the relative importance of the KL divergence term, ensuring that the learned latent space remains well-structured.

1. Mean Squared Error (MSE) Loss:
- The MSE Loss penalizes the squared difference between the predicted and ground-truth joint positions:

$$L_{MSE} = \sum_{i=1}^{N} M_i \cdot \left( \left( x_i^{recon} - x_i^{gt} \right)^2 + \left( y_i^{recon} - y_i^{gt} \right)^2 \right) \quad (2)$$

- This loss function is sensitive to outliers but provides a strong penalty for large deviations.

2. L1 Loss:
- The L1 Loss penalizes the absolute difference between the predicted and ground-truth joint positions:

$$L_{L1} = \sum_{i=1}^{N} M_i \cdot \left( \left| x_i^{recon} - x_i^{gt} \right| + \left| y_i^{recon} - y_i^{gt} \right| \right) \quad (3)$$

- L1 Loss is less sensitive to outliers and tends to produce sparser solutions, which can be beneficial in the presence of noise.

3. KL Divergence Loss:
- The KL divergence term regularizes the latent space by minimizing the divergence between the learned latent distribution and the prior distribution:

$$L_{KL} = -\frac{1}{2} \sum_{i=1}^{D} (1 + \log(\sigma_i^2) - \mu_i^2 - \sigma_i^2) \quad (4)$$

These loss formulations allow us to investigate the impact of different reconstruction loss functions on the overall performance of the DanceFusion framework, particularly in scenarios with incomplete or noisy data.

### 3.4. Handling Incomplete Skeleton Data

One of the major challenges in motion reconstruction is handling incomplete skeleton data, where certain joints may be missing due to occlusions or sensor noise. In the DanceFusion framework, this is addressed through a masking mechanism applied during the encoding phase. Each joint in the input sequence is either present or missing, and this information is captured in a binary mask. The mask prevents the model from considering missing joints during the spatial encoding process, ensuring that only reliable joints contribute to the final representation.

Formally, let $X \in R^{T \times J \times D}$ represent the input skeleton sequence, where $T$ is the number of frames, $J$ is the number of joints, and $D$ is the dimensionality of the joint position (e.g., 2D or 3D coordinates). The corresponding confidence matrix $M \in \{0,1\}^{T \times J}$ is defined as:

$$M_{t,j} = \begin{cases} 1 & \text{if joint j in frame t is present} \\ 0 & \text{if joint j in frame t is missing} \end{cases} \quad (5)$$

The masking mechanism modifies the input skeleton sequence by applying the confidence matrix $M$:

$$\widetilde{X}_{t,j} = M_{t,j} \cdot X_{t,j} \quad (6)$$

This masked sequence $\widetilde{X}$ is then fed into the encoder, ensuring that missing joints do not contribute to the reconstruction loss. The application of the mask in the computation of the loss function prevents the model from being penalized for missing data, allowing it to focus on accurately reconstructing the available joints.

## 3.5. Diffusion Model for Audio-Driven Motion Generation

Diffusion models have gained prominence in the generation of complex data sequences, demonstrating exceptional ability in synthesizing realistic and contextually accurate motion sequences. In DanceFusion, we leverage diffusion models to refine skeleton sequences and synchronize dance movements with audio inputs, thus enhancing both the visual quality and the auditory alignment of generated dance sequences.

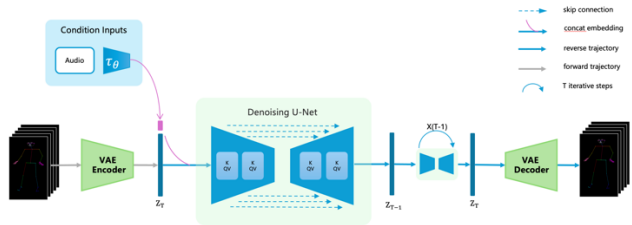

Figure 3: Diffusion Process.

### 3.5.1 Diffusion Process in Skeleton Sequence Refinement

Starting with an initial noisy skeleton sequence, the diffusion model iteratively applies transformations that progressively reduce noise and enhance the fidelity of the generated motion. This process ensures that the final output is not only realistic but also temporally consistent with the input dance or audio context.

The diffusion process is represented mathematically as follows:

$$L_{\text{diffusion}} = \mathbb{E}_{z_0, z_1, \ldots, z_T} \left[ \sum_{t=0}^{T-1} | z_{t+1} - z_t + \epsilon_t(z_t, \theta) |^2 \right] \quad (7)$$

Where $z_t$ represents the latent space at time step $t$, and $\epsilon_t$ denotes the noise model applied at each step, focusing on refining spatial accuracy and coherence without external audio cues.

### 3.5.2 Audio Feature Extraction

Audio features are extracted using a pre-trained model that captures the rhythmic and melodic characteristics of the input audio. These features are then utilized as conditional inputs to guide the diffusion process, enabling the generation of dance motions that are in sync with the audio.

Audio feature extraction is performed using Librosa[36], where 35-dimensional music temporal features are extracted following the FineDance[11] methodology. Additionally, a mel-spectrogram of the music clip is extracted and used as input for further processing.

Formally, let $x_{\text{audio}}$ represent the extracted audio features, and $z_0$ the initial latent vector for the motion sequence. The model is conditioned on these features as follows:

$$z_{t+1} = z_t - \epsilon_t(z_t, x_{\text{audio}}, \theta) \quad (8)$$

This conditioning ensures that each denoising step is influenced by the audio characteristics, aligning the generated motion with the rhythm and melody of the music.

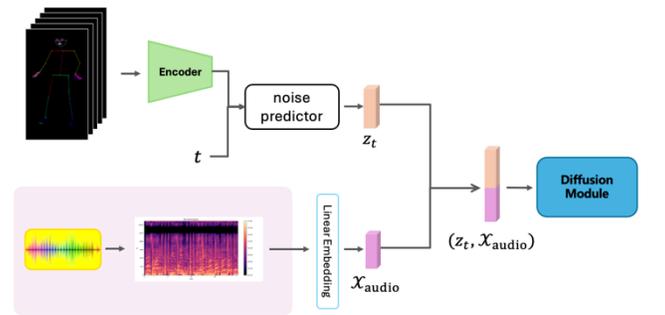

Figure 4: Diagram of the audio feature extraction process and its integration into the diffusion model.

### 3.5.3 Diffusion Process Integration with Audio Features

The diffusion model also plays a critical role in generating motion sequences that are driven by audio inputs. By integrating the audio features extracted from the input track, the model can generate synchronized dance motions that reflect the rhythm, melody, and overall mood of the accompanying music.

The diffusion model is trained to iteratively refine a sequence of random noise into a plausible motion sequence, conditioned on both the latent variables from the VAE and the extracted audio features. The integration of the diffusion model allows for the generation of realistic and synchronized dance motions, even when the input skeleton data is incomplete or noisy.

Formally, the diffusion process for audio-driven motion generation can be expressed as:

$$L_{\text{audio-diffusion}} = E_{z_0, z_1, \ldots, z_T} \left[ \sum_{t=0}^{T-1} |z_{t+1} - z_t + \epsilon_t(z_t, x_{\text{audio}}, \theta)|^2 \right] \quad (9)$$

This loss function ensures that the generated sequences are not only realistic but also synchronized with the audio.

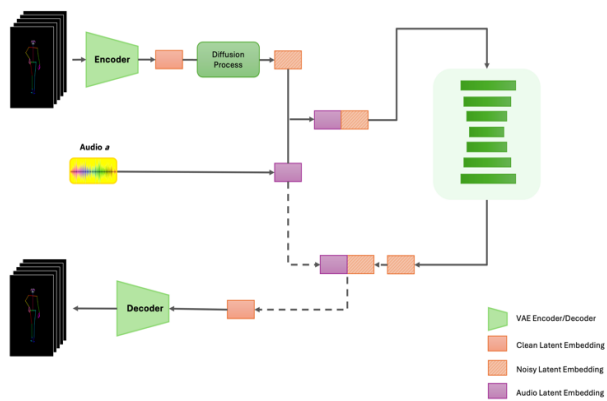

Figure 5: Visualization of the audio-driven diffusion process, highlighting the evolution of the motion sequence in synchronization with the audio.

### 3.6. Training Strategy

This section outlines the approach to optimizing the DanceFusion framework. It includes several key processes like data preprocessing, model training, and the optimization strategies applied during the learning phase.

1. **Data Preprocessing**: Skeleton data from TikTok videos are processed to extract meaningful sequences, such as identifying missing joints and synchronizing the skeleton data with corresponding audio features. This ensures that the input skeleton sequences are temporally aligned with the audio.
2. **Model Training**: The training phase includes a combination of reconstruction loss and KL-divergence loss. It incorporates masking techniques for handling missing joints and uses the AdamW optimizer for parameter updates. The learning rate schedule is fine-tuned to ensure a smooth convergence of the model.
3. **Loss Functions**: The loss function used in the training combines the reconstruction loss (MSE or L1) and the Kullback-Leibler divergence to ensure the stability of the latent space. The masking strategy during the training phase enhances the model's ability to handle incomplete skeleton data.
4. **Optimization Strategy**: This strategy includes the use of the AdamW optimizer, with learning rate schedules adjusted for optimal convergence over time, reducing the risk of overfitting.

### 3.7. Advantages of DanceFusion in Skeleton-Based Motion Analysis

This section highlights the key strengths of DanceFusion in skeleton-based motion analysis:
- Robust Handling of Incomplete Data: The integration of diffusion models enables DanceFusion to effectively deal with noisy and incomplete skeleton data, a common challenge in social media videos.
- High-Quality Motion Generation: By combining spatio-temporal encoding with diffusion-based refinement, DanceFusion produces smooth, realistic, and rhythmically accurate dance sequences.
- Scalability: The framework is flexible enough to be applied to different dance styles, making it versatile for various skeleton-based motion tasks.

## 4. Experimental Setup

This chapter details the experimental setup used to validate the performance of the DanceFusion framework, including a description of the datasets, evaluation metrics, and implementation specifics.

### 4.1. Dataset Description

The experiments utilize a custom-collected TikTok dance dataset comprising over 3,000 sequences, showcasing a diverse range of styles and rhythms typical of TikTok videos. This dataset includes multiple sequences of popular dance moves performed by various individuals, reflecting a wide range of styles and rhythms common in TikTok videos. Each dance sequence is processed into skeleton joint sequences, where each frame contains 137 joint coordinates. To ensure the model's generalization ability, the data is balanced across different dancers, dance styles, and various camera angles.

Each dance sequence is accompanied by a corresponding audio file that captures the rhythm and melody of the background music. These audio files are crucial for the audio-driven dance generation aspect of the framework.

This dataset is particularly challenging due to the presence of missing joints and noisy data, which is common in user-generated content. Our framework is designed to handle such variability effectively through the use of diffusion models and spatial-temporal encoding.

### 4.2. Evaluation Metrics

The performance of the DanceFusion framework is evaluated using the following metrics:

1. Fréchet Inception Distance (FID):

FID is used to assess the quality of the generated dance sequences by comparing the distribution of real dance sequences with the distribution of generated ones. A lower FID score indicates that the generated sequences are more realistic and closer to the real data distribution.

$$\text{FID} = \left\| \mu_r - \mu_g \right\|^2 + \text{Tr}\left(\Sigma_r + \Sigma_g - 2(\Sigma_r \Sigma_g)^{\frac{1}{2}}\right) \quad (10)$$

where $\mu_r$ and $\mu_g$ are the mean feature vectors, and $\Sigma_r$ and $\Sigma_g$ are the covariance matrices of the real and generated data distributions, respectively.

2. Diversity:

Diversity measures the variation within the generated dance sequences. It is crucial to ensure that the generated motions do not become repetitive and that they capture the range of styles and movements present in the dataset. We compute diversity as the variance across joint positions over different generated sequences.

Equation for Diversity:

$$\text{Diversity} = \frac{1}{N} \sum_{i=1}^{N} \text{Var}(X_i) \quad (11)$$

Where $X_i$ represents the joint positions for sequence i, and N is the number of sequences.

### 4.3. Implementation Details

#### 4.3.1 Hardware and Software Environment

All experiments are conducted on a workstation equipped with 8 NVIDIA H800 GPUs. The model is implemented using the PyTorch deep learning framework, with PyTorch Lightning used to streamline the training process. Additionally, the LibROSA library is employed for audio feature extraction.

#### 4.3.2 Hyperparameter Settings

The following hyperparameters are used for training the model:

- Batch Size: 1
- Learning Rate: 1e-5
- Loss function: A combination of the reconstruction loss (MSE or L1) and KL divergence, as described in Section 3.3.
- Sequence Length:
    Primary Setting: 30 frames
    Extended Setting: 300 frames (for specific experiments)
- Latent Dimension: 768
- Number of Transformer Layers: 8 layers for spatial encoding and 8 layers for temporal encoding

Due to the substantial computational resources required for processing long sequences, the majority of our experiments were conducted using a sequence length of 30 frames. This length was chosen as it balances computational efficiency with the ability to capture essential temporal dynamics of dance motions. Training with 30-frame sequences allowed us to perform extensive hyperparameter tuning and evaluate multiple configurations (e.g., with or without masking, MSE vs. L1 loss functions) within a reasonable time frame.

To assess the model's capability in handling longer and more complex motion sequences, we also trained a single configuration using a sequence length of 300 frames. This configuration employed masking and utilized the L1 loss function. Due to the increased computational demands—significantly longer training times and higher memory usage—we limited the 300-frame training to this specific setup. The purpose of this extended sequence training was primarily for demonstrative purposes, showcasing the model's potential in reconstructing and generating longer dance sequences with improved temporal coherence.

These hyperparameters were selected and refined through iterative experimentation to optimize the model's performance in both reconstruction and generation tasks. The combination of shorter and longer sequence lengths provided a comprehensive evaluation of the framework's capabilities across different temporal scales.

#### 4.3.3 Training Procedure

The training of the DanceFusion framework is divided into two main stages:

1. **VAE Pre-training**: Initially, the VAE component is pre-trained to accurately reconstruct the input skeleton data. The primary objective of this stage is to optimize the parameters of the spatial and temporal encoders, enabling the model to generate realistic joint positions from the latent space.
2. **Audio-Driven Generation Training**: Following the pre-training, the diffusion model is trained with the integration of audio features. The focus of this stage is to ensure that the generated dance motions are synchronized with the input audio, producing coherent and rhythmically accurate dance sequences.

The entire training process spans approximately 5000 epochs, with model checkpoints saved every 100 epochs for subsequent analysis.

## 5. Results and Discussion

This chapter evaluates the performance of the DanceFusion framework based on key metrics: **Fréchet Inception Distance (FID)** for reconstruction realism and **Diversity Score** for generated motion variety. We also investigate the model's robustness to incomplete skeleton data. The results are presented for different configurations: MSE vs. L1 loss functions, with and without masking.

### 5.1. Motion Reconstruction Performance

The **motion reconstruction performance** is crucial for assessing how well the DanceFusion framework can recover motion from incomplete or noisy skeleton data. The **FID** metric is used as the primary indicator of reconstruction quality, comparing generated skeleton motions with real TikTok dance data.

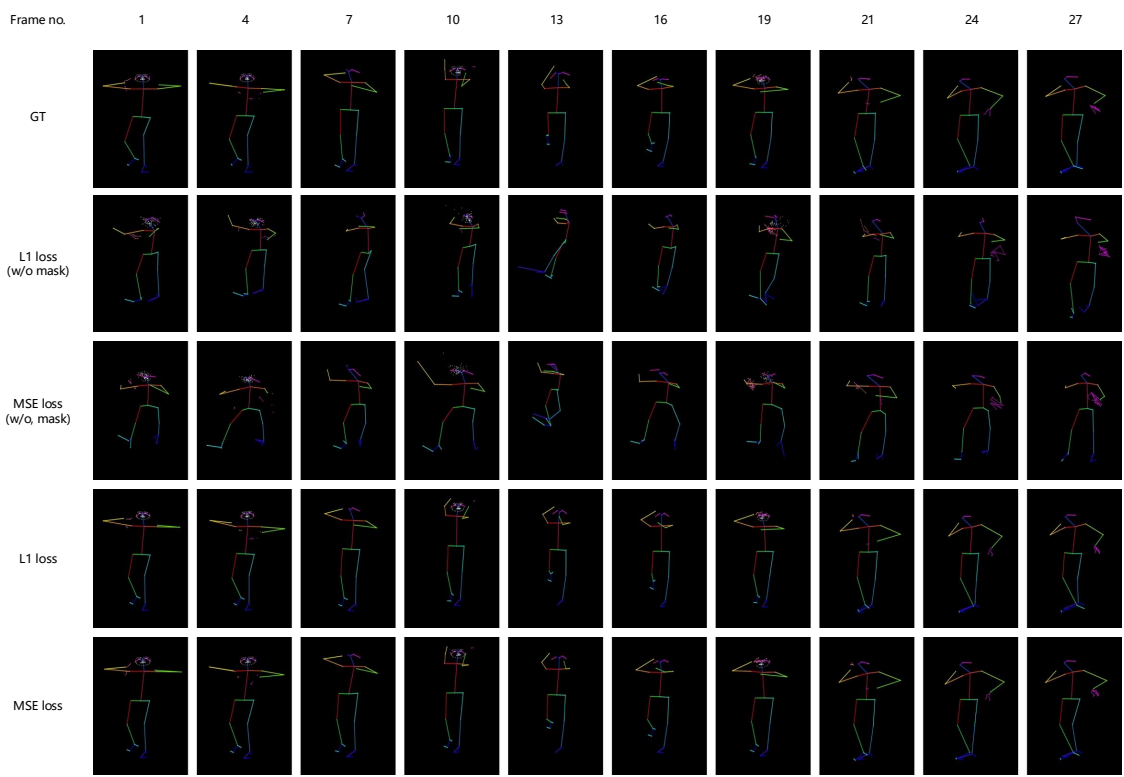

(a)

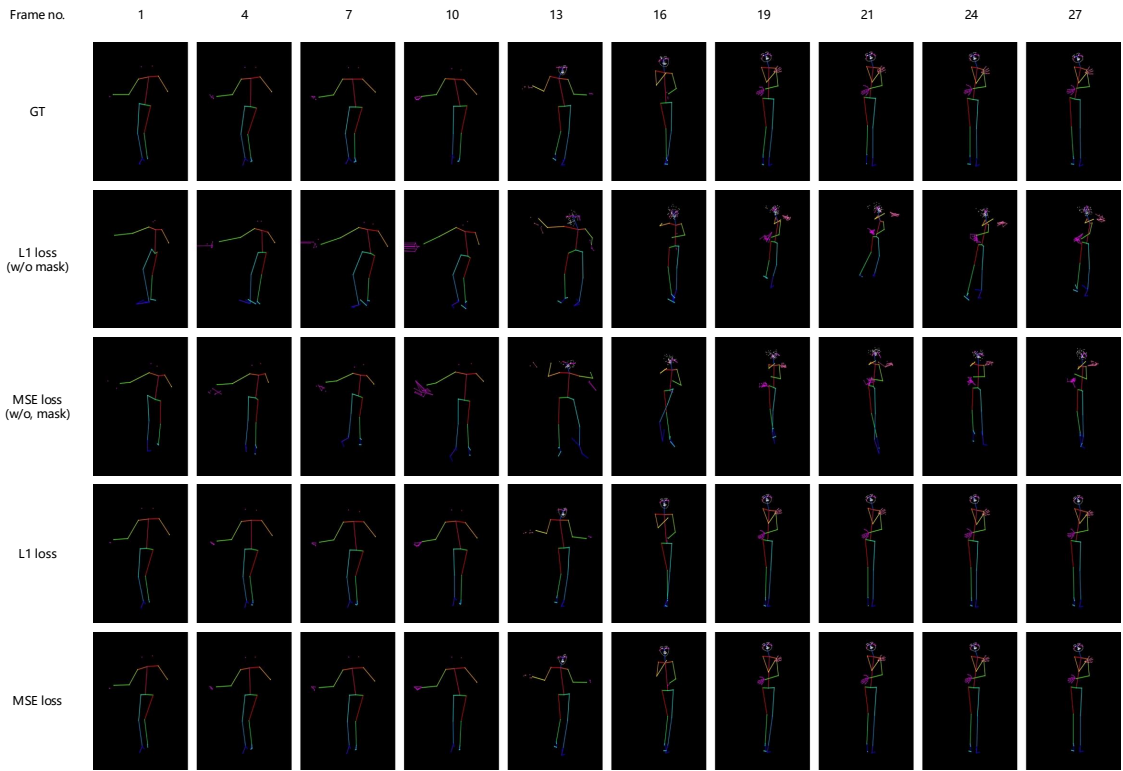

(b)

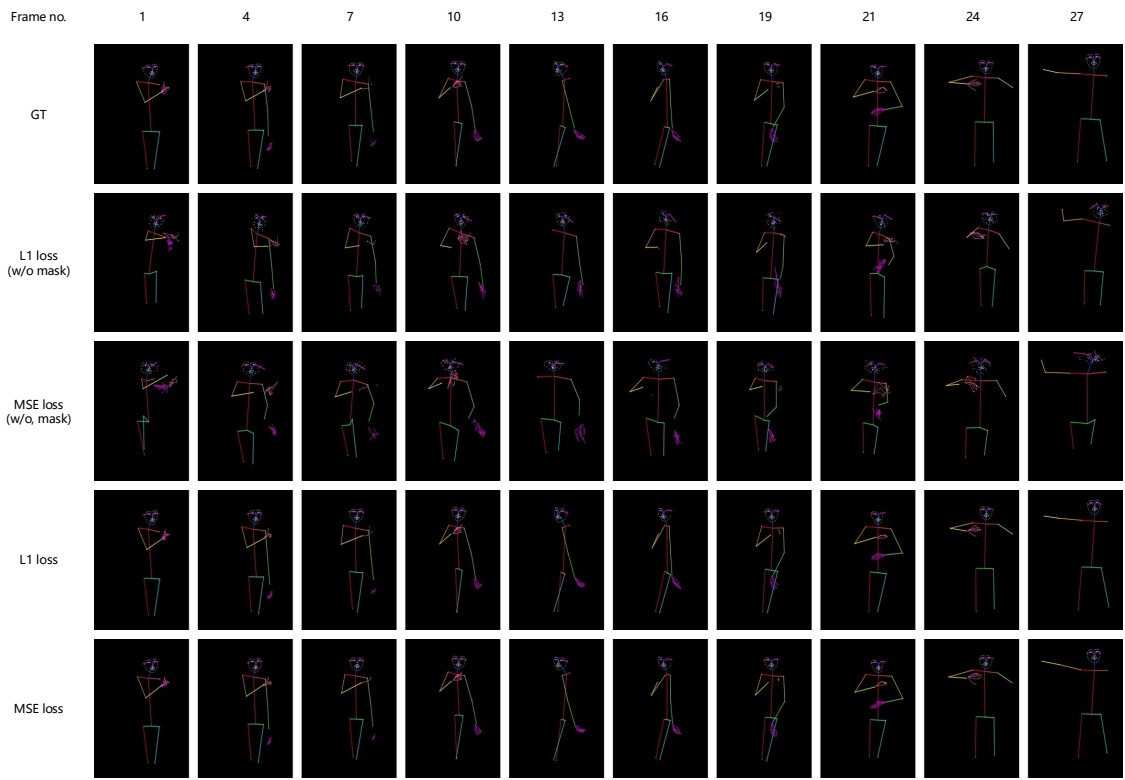

(c)

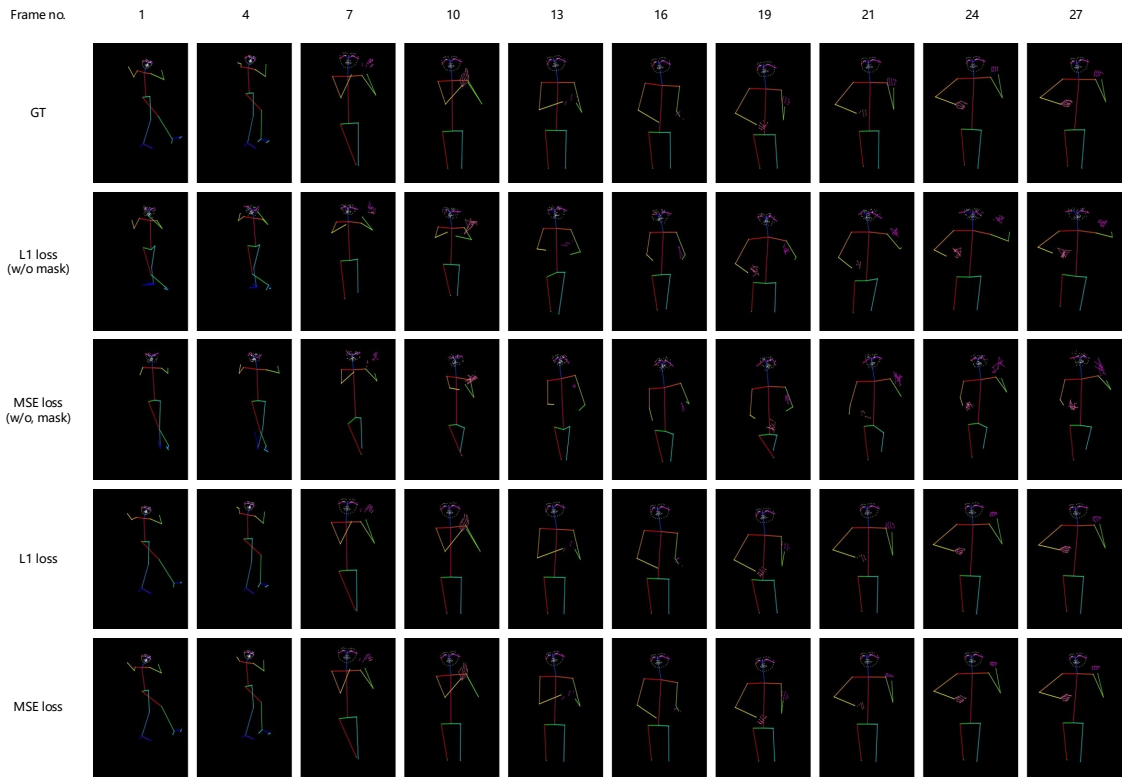

(d)

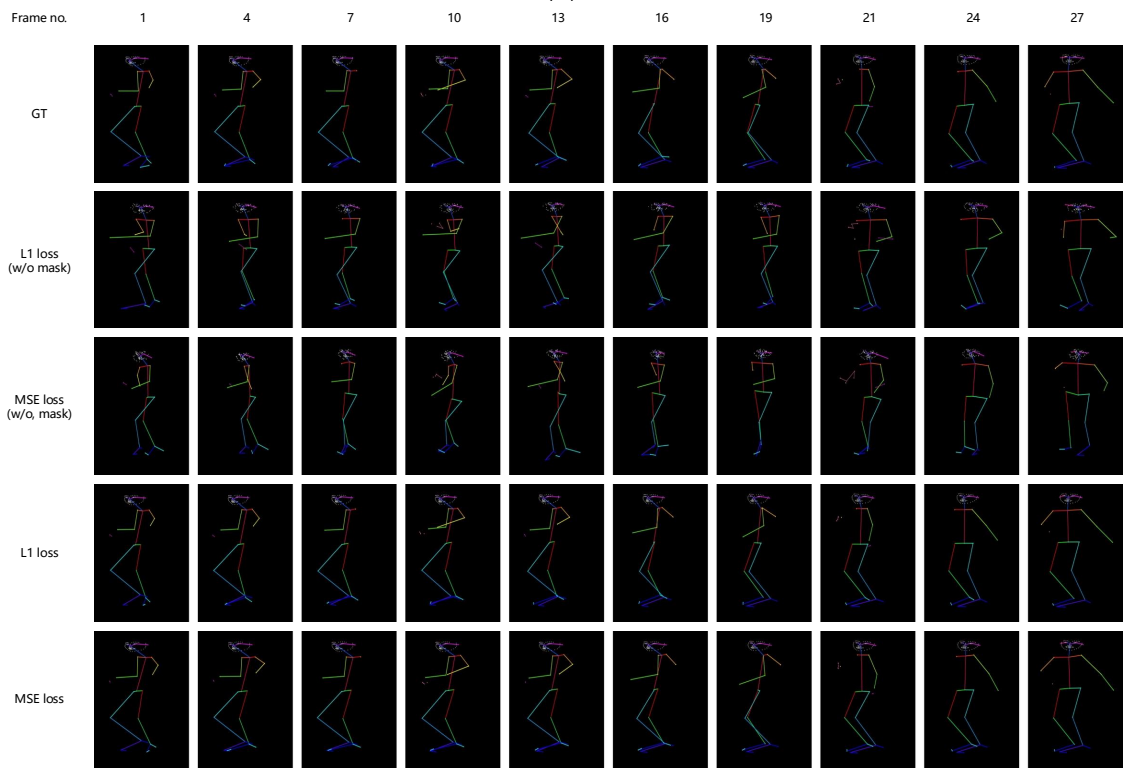

(e)

Figure 6: Comparisons of Reconstructed Motion Sequences using L1 Loss and MSE Loss with and without Masking. This figure showcases the reconstructed motion sequences for four configurations: MSE loss without masking, L1 loss without masking, MSE loss with masking, and L1 loss with masking, across the same input sequence.

### 5.1.1 Effectiveness of Masking

In this section, we evaluate the effectiveness of the masking technique in handling missing joints, using both the TikTok dance dataset and the AIST dataset. Unlike TikTok data, which often contains incomplete skeletons due to occlusions and sensor limitations, the AIST dataset provides fully intact skeletal sequences without missing joints. Therefore, the experiments on the AIST dataset do not include "without mask" configurations, as masking is unnecessary for fully complete sequences.

As shown in Table 1, models trained on the AIST++[12] dataset achieve low FID scores for both MSE and L1 loss functions, indicating robust performance in reconstructing high-quality motion sequences from complete data. This result demonstrates DanceFusion's ability to perform accurate reconstructions when provided with clean, fully observed skeletal data.

In contrast, for the TikTok dataset, where missing joints are prevalent, the use of masking significantly improves the FID scores across both MSE and L1 loss configurations. This highlights the importance of masking in handling real-world, noisy data. The difference in performance between the AIST++ and TikTok datasets underscores DanceFusion's adaptability and the critical role of masking when reconstructing sequences with incomplete data.

| Model Configuration | FID ↓ |
|---|---|
| AIST++ [12] - MSE loss | 0.0011 |
| AIST++ [12] - L1 loss | 0.0025 |
| DanceFusion - MSE loss (w/o mask) | 10.0899 |
| DanceFusion - L1 loss (w/o mask) | 8.6359 |
| DanceFusion - MSE loss | 0.2344 |
| Dance Fusion - L1 loss | 0.1170 |

Table 1: FID Scores for Different Loss Functions and Configurations.

### 5.1.2 Comparison of Loss Functions

While both **MSE and L1 loss** functions produced similar results, the presence of masking had a far more substantial impact on performance. Table 1 shows minimal differences between MSE and L1 loss in the masked condition, suggesting that the key to reconstruction quality lies not in the choice of the loss function but in how missing data is handled.

Thus, we conclude that **masking is the critical factor** for enhancing motion reconstruction, especially when dealing with noisy or incomplete data. This finding is consistent across the experiments.

## 5.2. Diversity in Audio-Driven Motion Generation

Generating diverse, synchronized dance motions based on audio input is another critical aspect of the DanceFusion framework. Here, we evaluate the **Diversity Score**, which measures the variability of generated motion sequences for the same audio input.

### 5.2.1 Evaluation of Diversity

The Diversity Score captures how varied the generated dance motions are, even when the same audio track is used as input. Table 2 illustrates how different configurations of the model (MSE vs. L1 loss, with or without masking) perform in terms of diversity.

| Model Configuration | Diversity Score |
|---|---|
| Dance Fusion - L1 loss (w/o mask) | 7.4328 |
| DanceFusion - MSE loss (w/o mask) | 7.5255 |
| DanceFusion - L1 loss | 7.5482 |
| Dance Fusion - MSE loss | 7.6515 |

Table 2: Diversity Scores for Different Loss Functions and Configurations.

The results suggest that neither the choice of loss function (MSE vs. L1) nor the use of masking has a substantial impact on the diversity of generated motion sequences, as all configurations yield similar Diversity Scores. This indicates that the DanceFusion framework inherently possesses strong capability for generating varied motion sequences in response to the same audio input, independent of these specific configurations.

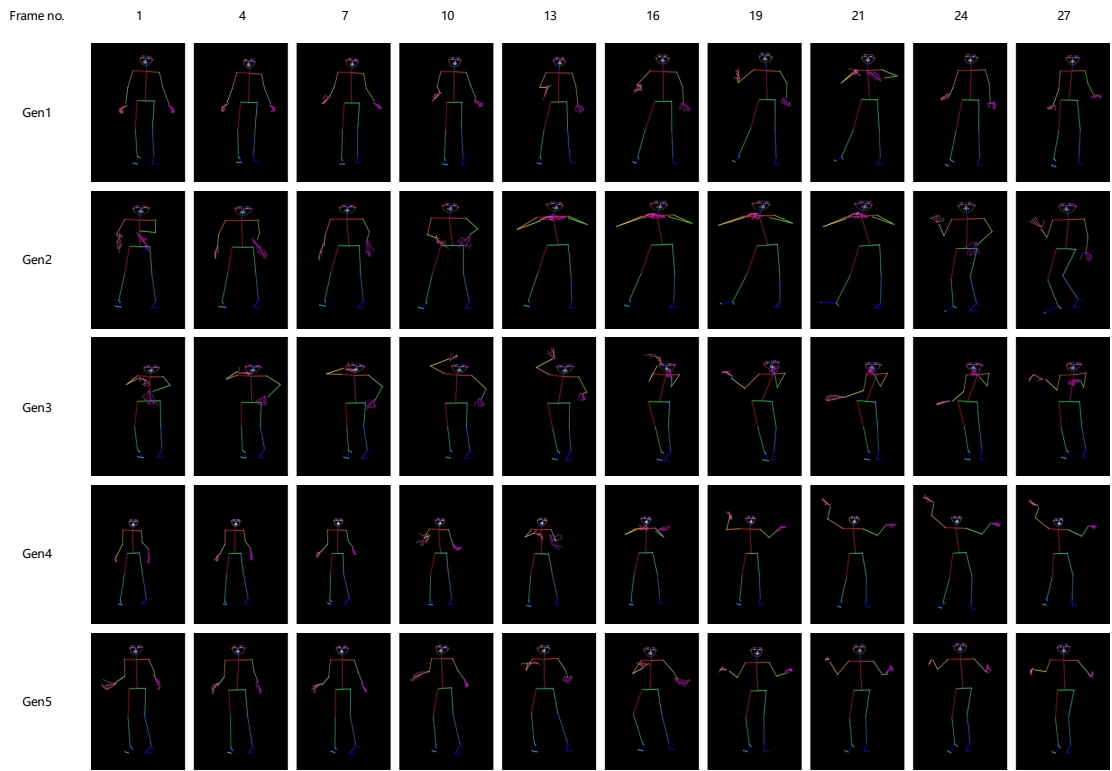
(music #1)

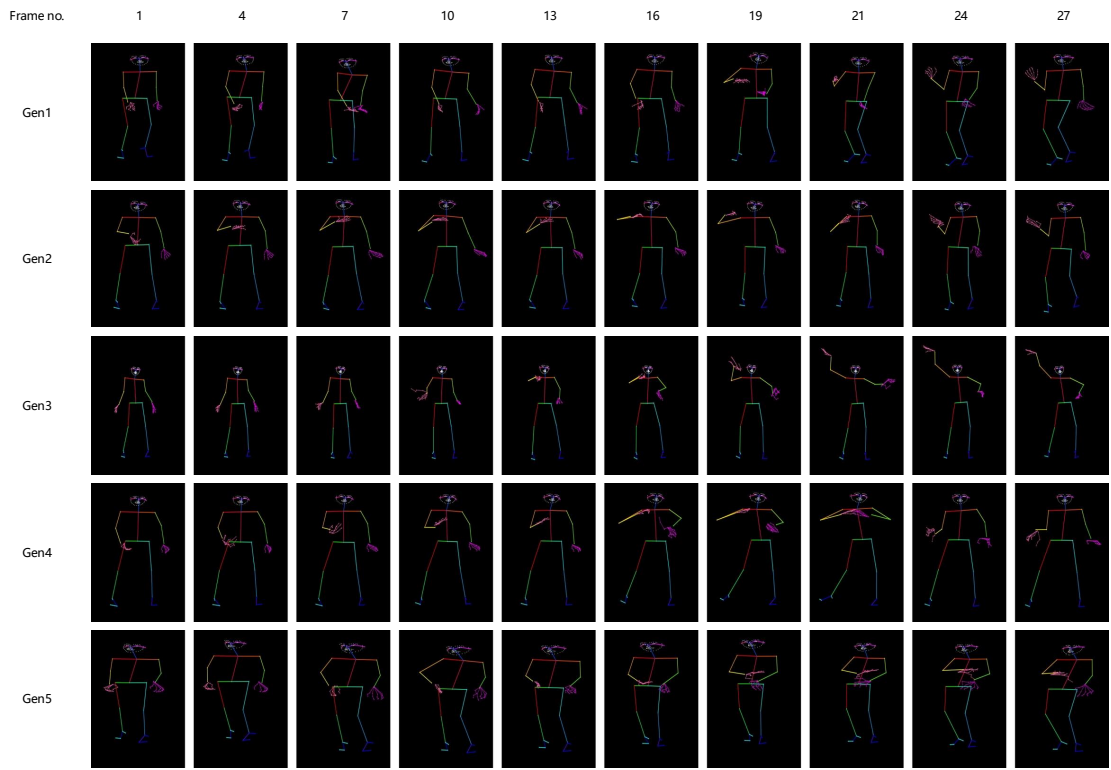
(music #2)

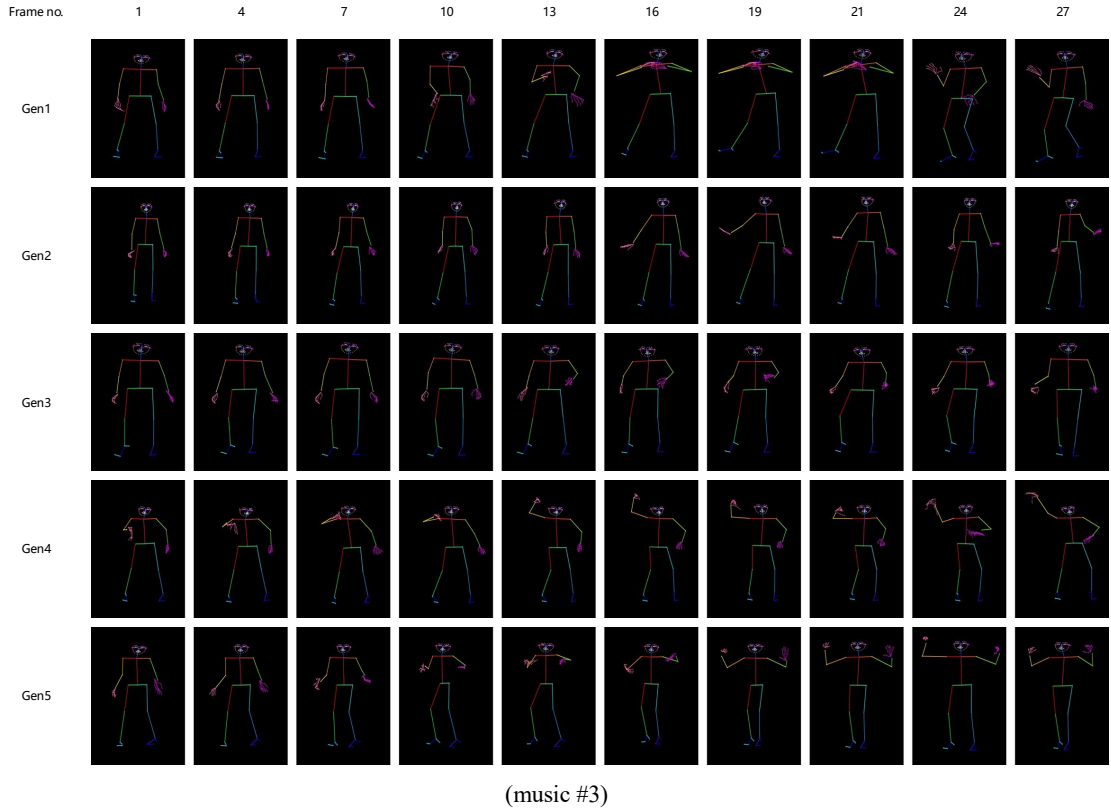

(music #3)

Figure 7: Examples of dance motion sequences generated by the DanceFusion framework from different music tracks.

## 5.3. Robustness to Incomplete Data

A crucial feature of the DanceFusion framework is its robustness to incomplete skeleton data, often encountered in real-world scenarios. The masking mechanism enables the model to focus on existing, reliable joints and disregard the missing ones, thereby reducing the negative impact of data incompleteness. Table 3 displays FID scores across various levels of missing joint data (5% to 20%), illustrating the significant improvement achieved by models with masking.

The FID scores in Table 3 indicate that models utilizing masking achieved significantly lower FID values across all missing data levels, showing improved reconstruction quality compared to models without masking. The model with masked L1 loss consistently outperformed other configurations, even as missing joint levels increased.

| Missing joint data | 5% ↓ | 10% ↓ | 15% ↓ | 20% ↓ |
|---|---|---|---|---|
| MSE loss (w/o mask) | 162.9972 | 416.0702 | 670.1728 | 886.4969 |
| L1 loss (w/o mask) | 109.7537 | 366.7350 | 779.8210 | 1147.7648 |
| MSE loss | 0.8086 | 1.2865 | 2.9940 | 3.7558 |
| L1 loss | 0.4084 | 1.7630 | 2.6089 | 2.7496 |

Table 3: FID for Different Levels of Missing Data

## 5.4. Summary of Key Findings

- Masking is Critical: Across all experiments, masking consistently improved reconstruction quality and motion diversity, as evidenced by lower FID and higher Diversity Scores. This technique allows the model to effectively handle missing joints, making it a crucial component for high-quality motion generation.
- Minimal Impact of Loss Function Choice: The difference between MSE and L1 losses was minor compared to the impact of masking, indicating that masking plays a more substantial role in enhancing model performance than loss function selection.
- Resilience to Data Incompleteness: The DanceFusion framework demonstrated robust performance even with significant levels of data incompleteness (up to 20%), largely due to the effective masking strategy.

These findings highlight the importance of robust data handling techniques, such as masking, and suggest that while the choice of loss function is less important, attention should be focused on improving the model's ability to deal with noisy or incomplete data.

## 6. Conclusion and Future Work

The DanceFusion framework represents a significant advancement in the field of motion generation and reconstruction, particularly in the context of social media-driven dance content like TikTok. By combining a hierarchical Transformer-based Variational Autoencoder (VAE) with diffusion models, this research has demonstrated a robust approach to handling incomplete skeleton data and generating realistic, synchronized dance sequences from audio inputs.

### 6.1. Summary of Key Findings

This study has successfully addressed several key challenges in motion generation and reconstruction:
- Hierarchical Spatio-Temporal Encoding: The integration of spatial and temporal encodings through hierarchical Transformers has enabled the model to capture intricate dance dynamics, ensuring robust motion reconstruction even from incomplete skeleton data.
- Diffusion-Based Motion Refinement: The introduction of diffusion models in refining motion sequences has proven effective in generating synchronized and realistic dance sequences, closely aligned with audio inputs.
- Masking Technique: The masking technique employed in the DanceFusion framework demonstrates significant improvements in handling incomplete or noisy skeleton data, which is essential for real-world applications.

### 6.2. Limitations

While the DanceFusion framework marks significant progress, several limitations remain:
1. **Data Dependency**: The model's performance is highly dependent on the quality and diversity of the training data. Although TikTok provides a vast amount of dance content, the informal and varied nature of these recordings can introduce noise and inconsistencies that may affect model performance.
2. **Computational Complexity**: The hierarchical Transformer and diffusion models are computationally intensive, requiring significant resources for training and inference. This complexity might limit the framework's scalability, especially for real-time applications or deployment on devices with limited processing power.
3. **Generalization to Other Dance Styles**: While the framework is effective for TikTok dances, which are typically short and rhythmically driven, its applicability to other types of dance or more complex motion sequences has not been extensively tested. Further research is needed to assess the model's generalization capabilities across different dance genres and motion contexts.

### 6.3. Future Research Directions

Building on the findings of this research, several avenues for future work are suggested:
1. **Efficiency Optimization**: Developing more efficient algorithms to reduce the computational complexity of both the hierarchical Transformer and diffusion processes will be critical, particularly for real-time motion generation applications.
2. **Multi-modal Integration**: Future work should explore incorporating additional modalities such as text, speech, or video into the framework. This could enable the generation of more nuanced and contextually appropriate dance sequences based on richer input data.
3. **Generalization Across Dance Styles**: Expanding the model to support a broader range of dance genres, including complex and extended motion sequences, will further demonstrate the versatility of the DanceFusion framework.

### 6.4. Final Remarks

The DanceFusion framework presents a step forward in the automatic generation and reconstruction of dance motions, addressing key challenges such as data incompleteness and synchronization with audio. The combined use of VAEs, Transformers, and diffusion models opens new possibilities for creative and technical advancements in motion analysis and generation. As the field evolves, further developments in efficiency, multimodal integration, and generalization will drive the next generation of motion generation systems for creative industries and beyond.

## 8. Appendices

### 8.1. Additional Experimental Results

To further illustrate the performance of the DanceFusion framework across different configurations and input scenarios, this section presents additional visualizations of reconstructed motion sequences and audio-driven generated dance motions.

#### 8.1.1 Additional Reconstructed Motion Sequences

In this subsection, we display a broader set of reconstructed motion sequences under varying configurations, including with and without masking, using both MSE and L1 loss functions. These additional examples serve to validate the consistency and robustness of the framework across multiple test samples. Each sequence is generated from ground truth skeleton data with varying levels of missing joints, highlighting the impact of masking in handling incomplete skeleton data.

#### 8.1.2 Additional Audio-Driven Generated Dance Motions

This subsection includes further examples of dance motions generated based on different music inputs. These examples illustrate the model's capability to produce diverse, synchronized motions that align with the rhythmic and stylistic characteristics of the audio input. Variability in generated motions across different audio inputs is demonstrated to emphasize the model's capacity for diversity in audio-driven dance generation.

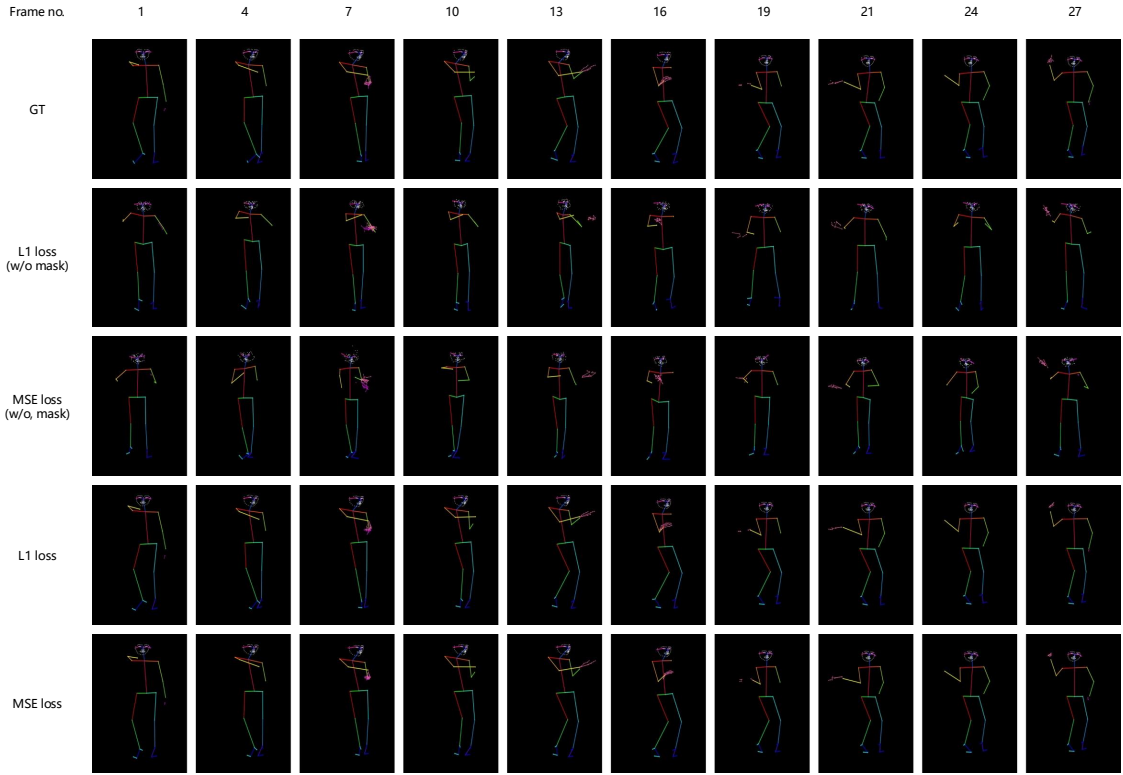

(a)

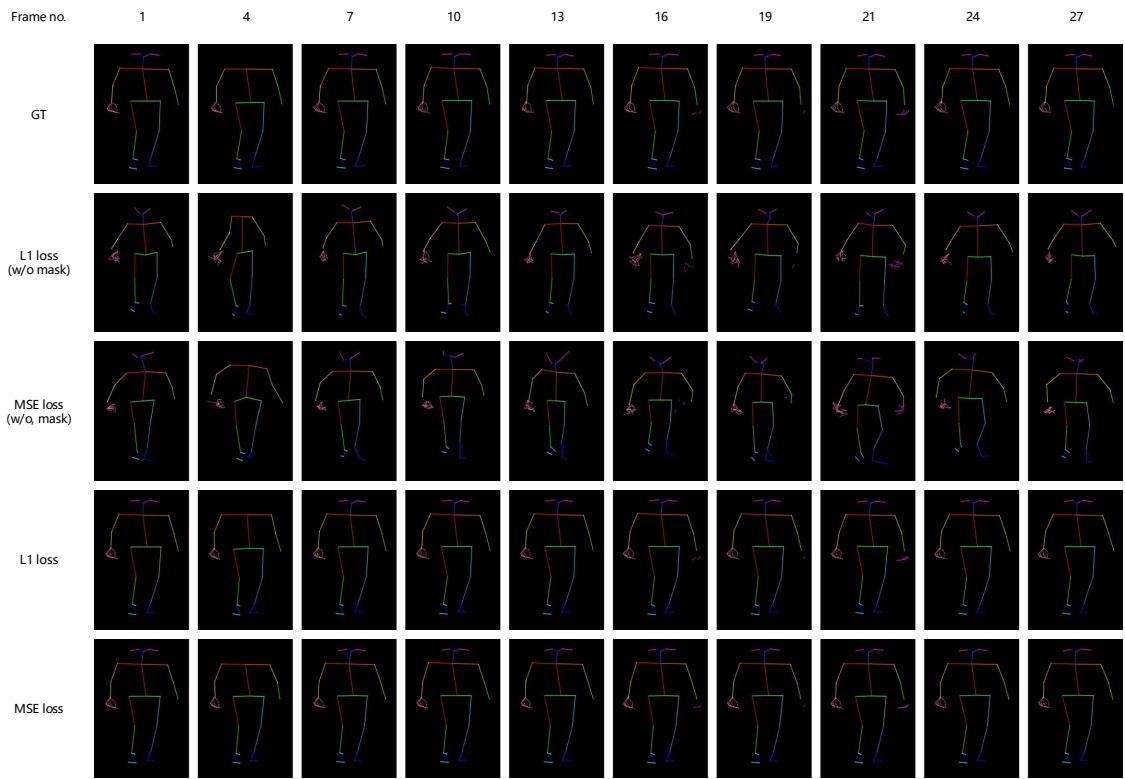

(b)

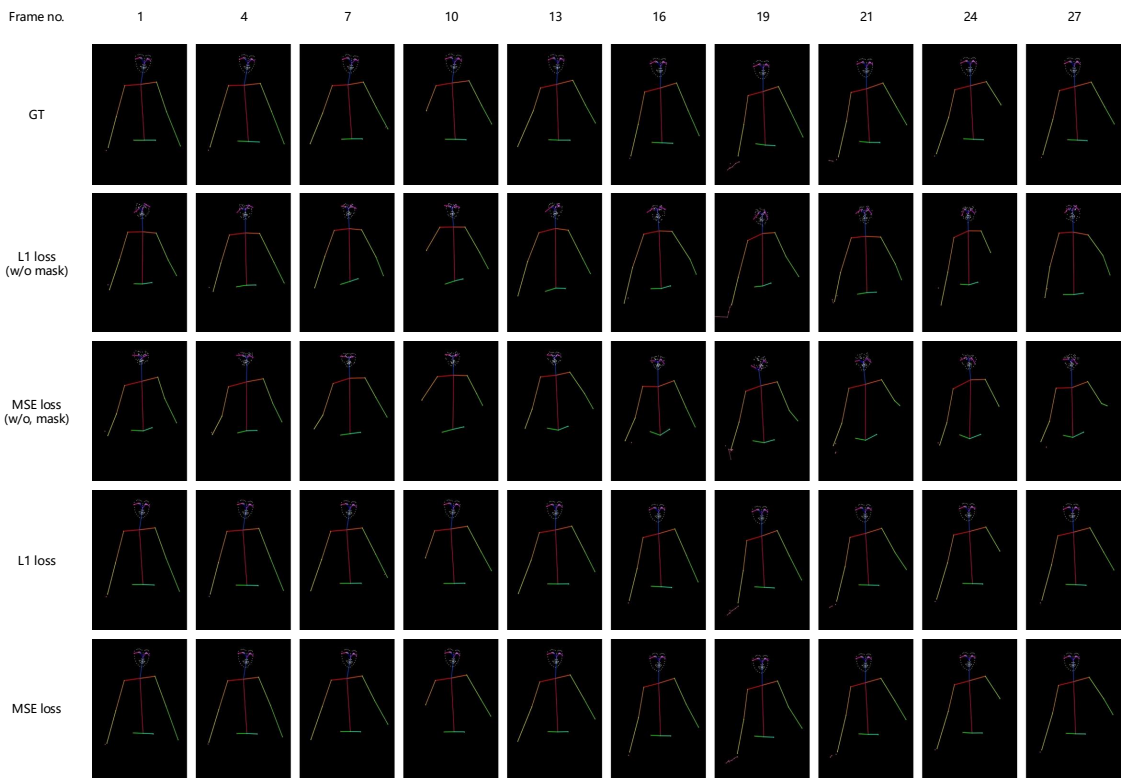

(c)

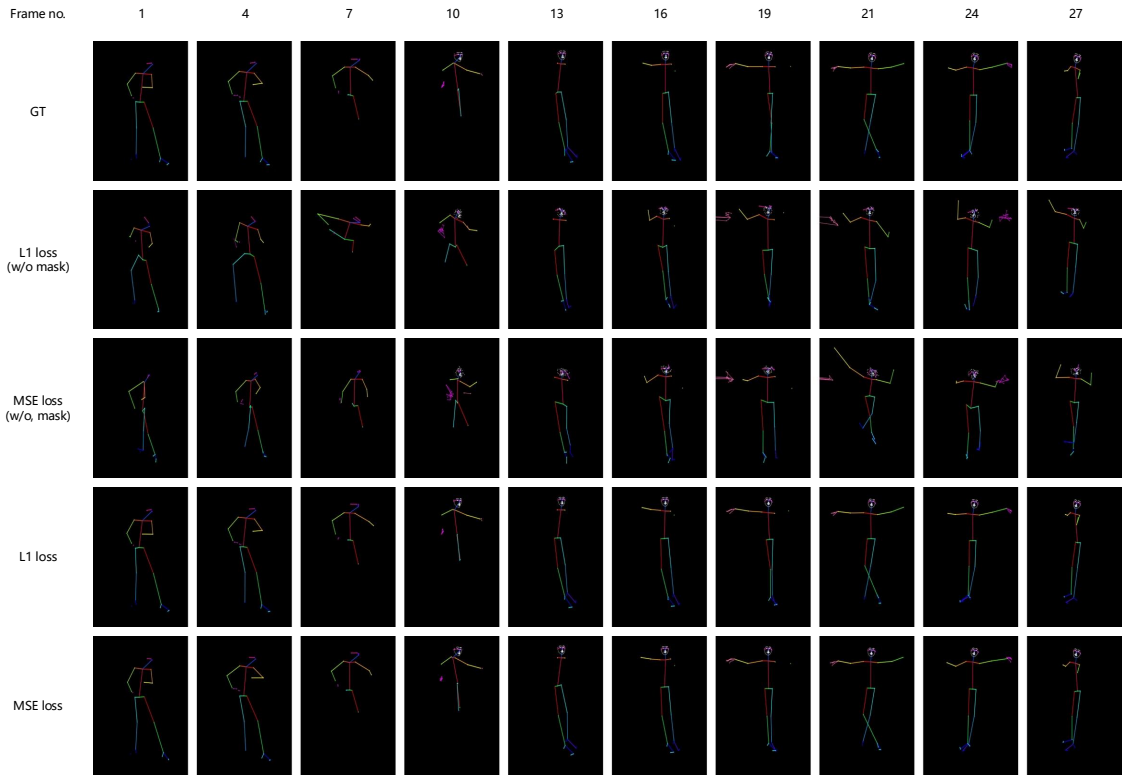

(d)

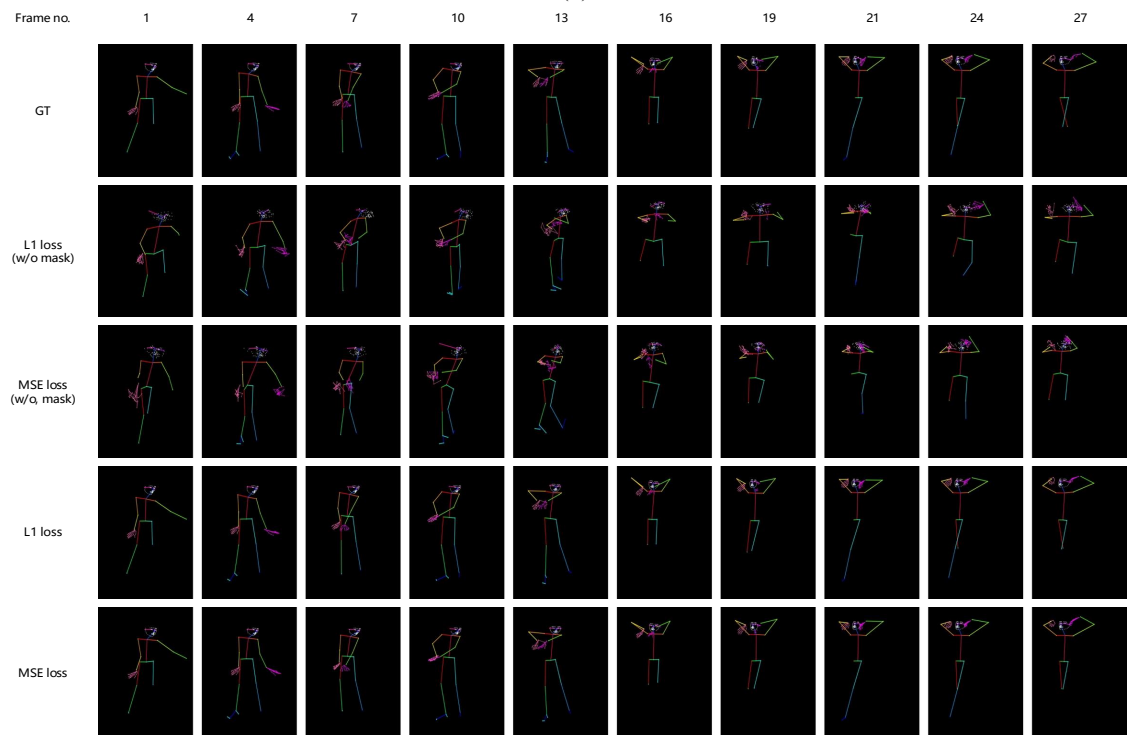

(e)

Figure 11: **Additional Comparisons of Reconstructed Motion Sequences using L1 Loss and MSE Loss with and without Masking.** This figure showcases the reconstructed motion sequences for four configurations: MSE loss without masking, L1 loss without masking, MSE loss with masking, and L1 loss with masking, across the same input sequence.

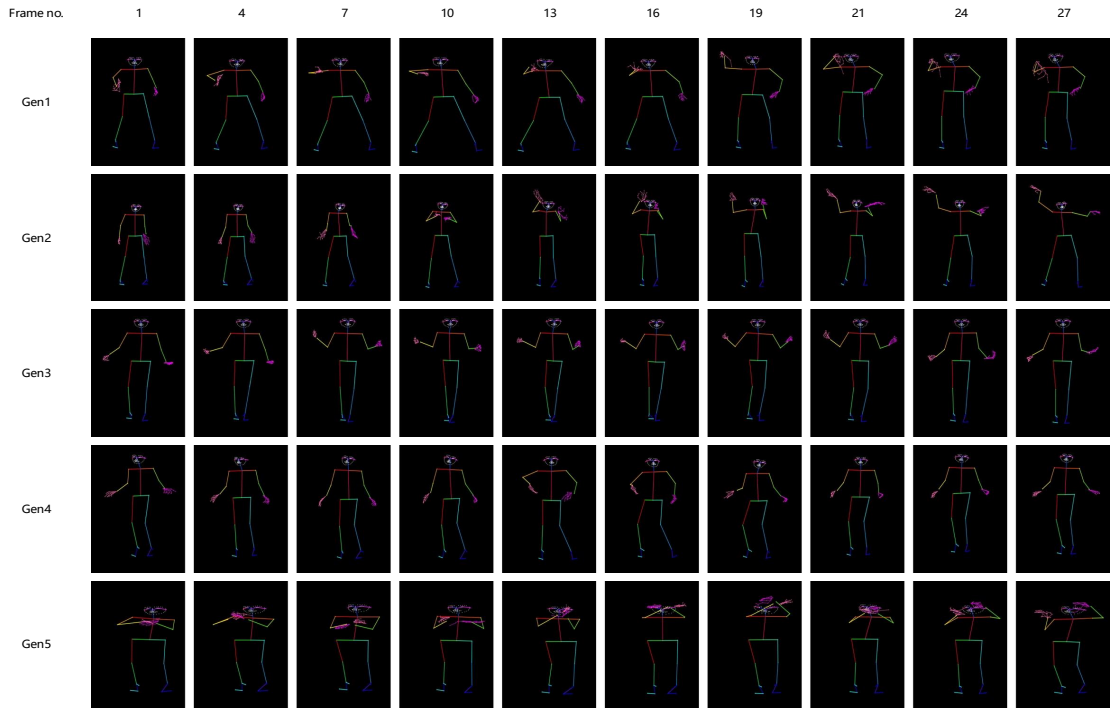

(music #1)

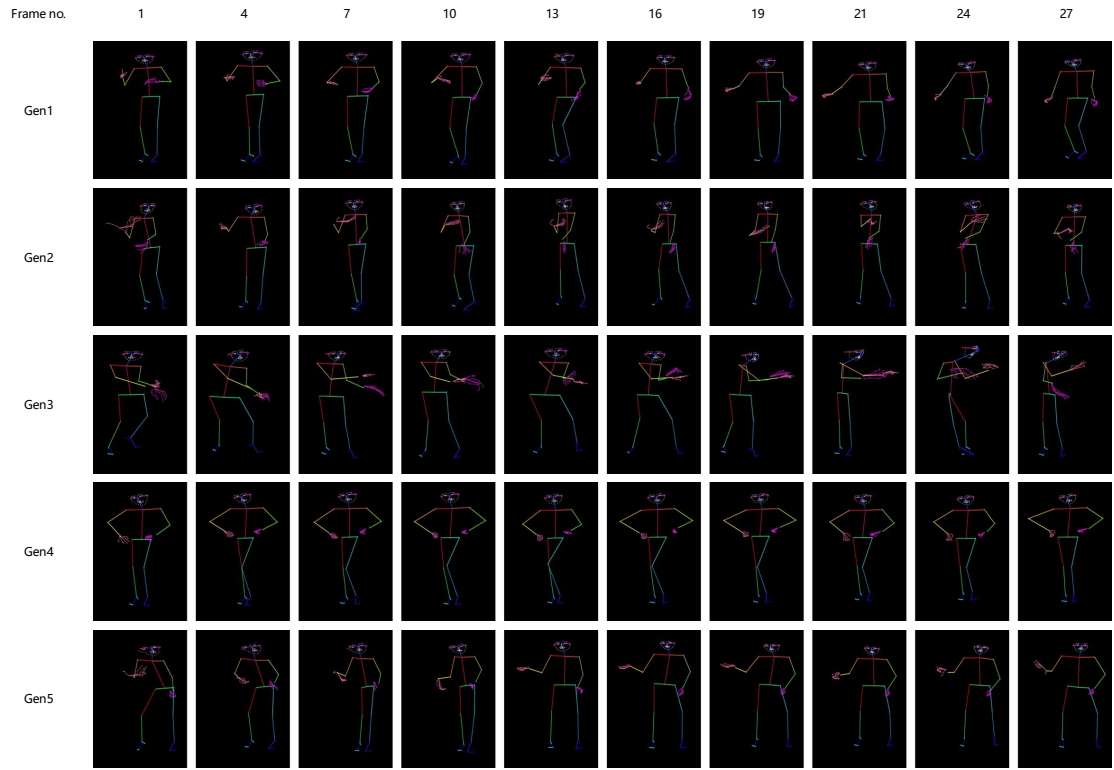

(music #2)

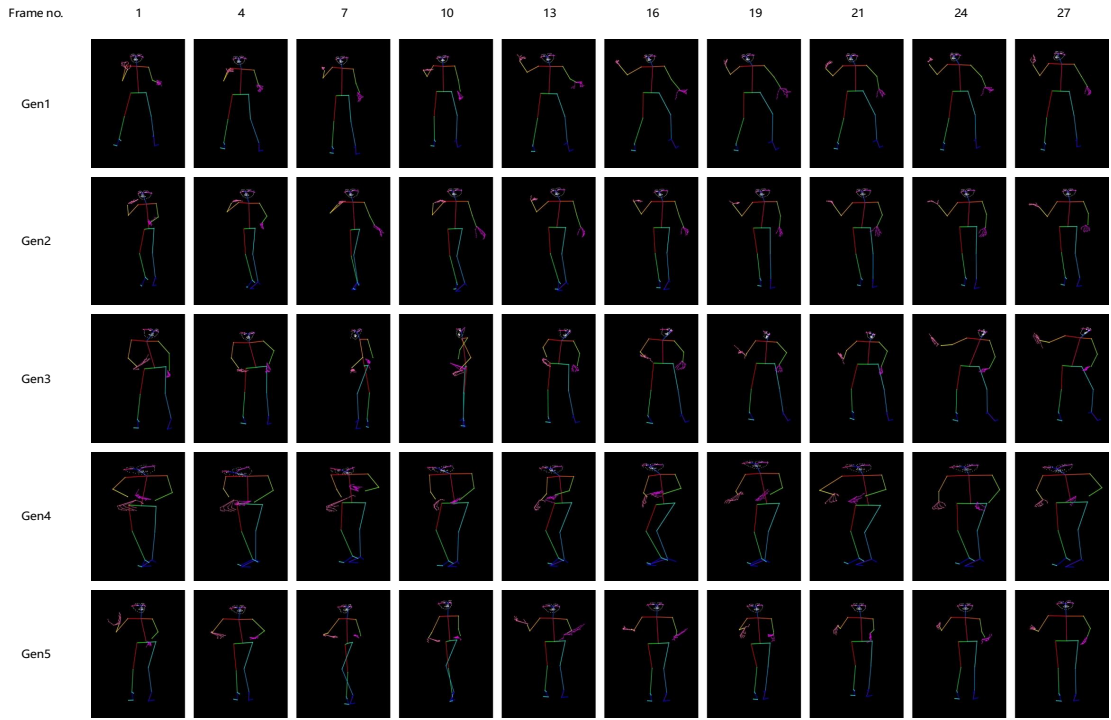

(music #3)

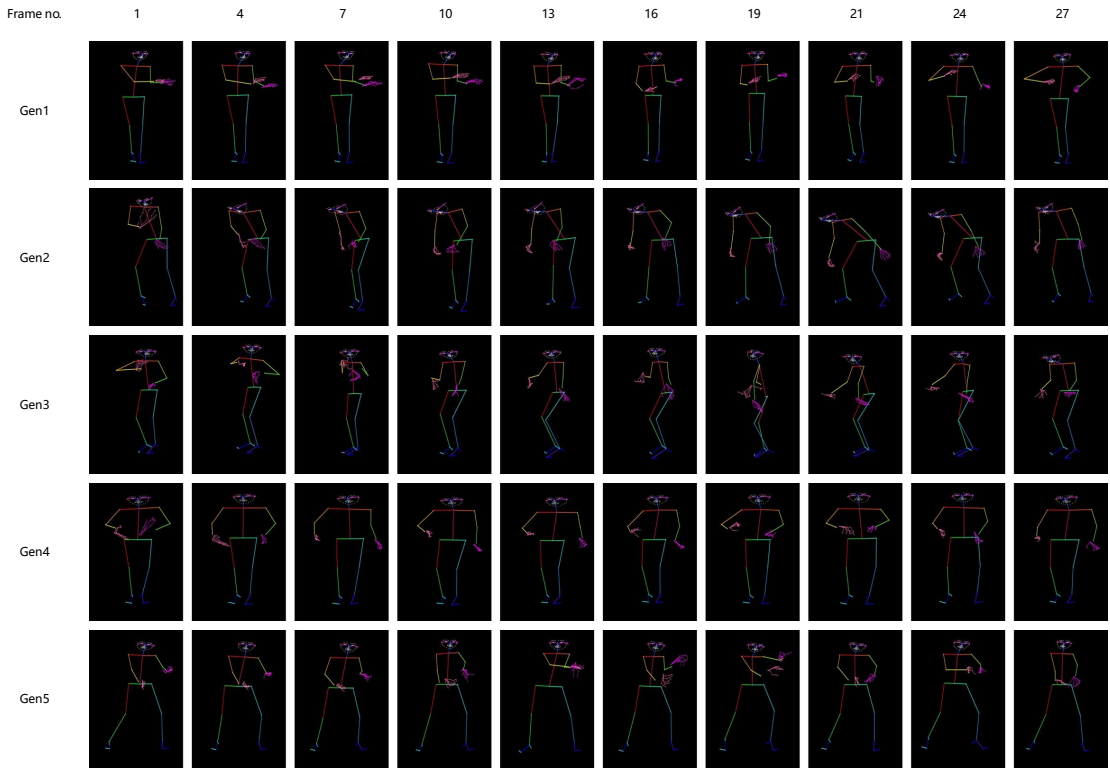

(music #4)

Figure 12: **Additional Audio-Driven Generated Dance Motions for Different Music Tracks.** This figure displays additional generated dance sequences synchronized with various music inputs, demonstrating the diversity and rhythmic alignment of the DanceFusion framework across different audio inputs.